\documentstyle[aclap]{article}

\title{\vspace{-0.5in}The TreeBanker: a Tool for \\ 
Supervised Training of Parsed Corpora}

\author{David Carter\\
SRI International \\ 
23 Millers Yard, Mill Lane\\
Cambridge CB2 1RQ\\
United Kingdom \\
{\tt dmc@cam.sri.com}}

\begin{document}
\bibliographystyle{fullname}

\maketitle
\vspace{-0.5in}

\begin{abstract}
I describe the TreeBanker, a graphical tool for the supervised
training involved in domain customization of the disambiguation
component of a speech- or language-understanding system. The
TreeBanker presents a user, who need not be a system expert, with a
range of properties that distinguish competing analyses for an
utterance and that are relatively easy to judge. This allows training
on a corpus to be completed in far less time, and with far less
expertise, than would be needed if analyses were inspected directly:
it becomes possible for a corpus of about 20,000 sentences of the
complexity of those in the ATIS corpus to be judged in around three
weeks of work by a linguistically aware non-expert.
\end{abstract}

\section{Introduction}

In a language understanding system where full,
linguistically-motivated analyses of utterances are desired, the
linguistic analyser needs to generate possible semantic
representations and then choose the one most likely to be correct. If
the analyser is a component of a pipelined speech understanding
system, the problem is magnified, as the speech recognizer will
typically deliver not a word string but an N-best list or a lattice;
the problem then becomes one of choosing between multiple analyses of
several competing word sequences.

In practice, we can only come near to satisfactory disambiguation
performance if the analyser is trained on a corpus of utterances from
the same source (domain and task) as those it is intended to
process. Since this needs to be done afresh for each new source, and
since a corpus of several thousand sentences will normally be needed,
economic considerations mean it is highly desirable to do it as
automatically as possible.  Furthermore, those aspects that cannot be
automated should as far as possible not depend on the attention of
experts in the system and in the representations it uses.

The Spoken Language Translator (SLT; Becket {\it et al}, forthcoming;
Rayner and Carter, 1996 and 1997) is a pipelined speech
understanding system of the type assumed here. It is constructed from
general-purpose speech recognition, language processing and speech
synthesis components in order to allow relatively straightforward
adaptation to new domains.  Linguistic processing in the SLT system is
carried out by the Core Language Engine (CLE; Alshawi, 1992). Given an
input string, N-best list or lattice, the CLE applies
unification-based syntactic rules and their corresponding semantic
rules to create zero or more quasi-logical form (QLF, described below;
Alshawi, 1992; Alshawi and Crouch, 1992) analyses of it;
disambiguation is then a matter of selecting the correct (or at least,
the best available) QLF.

This paper describes the TreeBanker, a program that facilitates
supervised training by interacting with a non-expert user and that
organizes the results of this training to provide the CLE with data in
an appropriate format. The CLE uses this data to analyse speech
recognizer output efficiently and to choose accurately among the
interpretations it creates.  I assume here that the coverage problem
has been solved to the extent that the system's grammar and lexicon
license the correct analyses of utterances often enough for practical
usefulness (Rayner, Bouillon and Carter, 1995). 

The examples given in this paper are taken from the ATIS (Air Travel
Inquiry System; Hemphill {\it et al}, 1990) domain. However, wider
domains, such as that represented in the North American Business News
(NAB) corpus, would present no particular problem to the TreeBanker as
long as the (highly non-trivial) coverage problems for those domains
were close enough to solution. The examples given here are in fact all
for English, but the TreeBanker has also successfully been used for
Swedish and French customizations of the CLE (Gamb\"{a}ck and Rayner,
1992; Rayner, Carter and Bouillon, 1996).

\section{Representational Issues}
\label{repissues}

In the version of QLF output by the CLE's analyser, content word
senses are represented as predicates and predicate-argument relations
are shown, so that selecting a single QLF during disambiguation
entails resolving content word senses and many structural ambiguities.
However, many function words, particularly prepositions, are not
resolved to senses, and quantifier scope and anaphoric references are
also left unresolved. Some syntactic information, such as number and
tense, is represented. Thus QLF encodes quite a wide range of the
syntactic and semantic information that can be useful both in
supervised training and in run-time disambiguation.

QLFs are designed to be appropriate for the inference or other
processing that follows utterance analysis in whatever application
(translation, database query, etc.) the CLE is being used
for. However, they are not easy for humans to work with directly in
supervised training. Even for an expert, inspecting all the analyses
produced for a sentence is a tedious and time-consuming task. There
may be dozens of analyses that are variations on a small number of
largely independent themes: choices of word sense, modifier
attachment, conjunction scope and so on. Further, if the
representation language is designed with semantic and computational
considerations in mind, there is no reason why it should be easy to
read even for someone who fully understands it. And indeed, as
already argued, it is preferable that selection of the correct
analysis should as far as possible not require the intervention of
experts at all. The TreeBanker (and, in fact, the CLE's preference
mechanism, omitted here for space reasons but discussed in detail by
Becket {\it et al}, forthcoming) therefore treats a QLF as completely
characterized by its {\it properties}: smaller pieces of information,
extracted from the QLF or the syntax tree associated with it, that are
likely to be easy for humans to work with.

The TreeBanker presents instances of many kinds of property to the
user during training.  However, its functionality in no way depends on
the specific nature of QLF, and in fact its first action in the
training process is to extract properties from QLFs and their
associated parse trees, and then never again to process the QLFs
directly.  The database of analysed sentences that it maintains
contains only these properties and not the analyses themselves.  It
would therefore be straightforward to adapt the TreeBanker to any
system or formalism from which properties could be derived that both
distinguished competing analyses and could be presented to a
non-expert user in a comprehensible way. Many mainstream systems and
formalisms would satisfy these criteria, including ones such as the
University of Pennsylvania Treebank (Marcus {\it et al}, 1993) which
are purely syntactic (though of course, only syntactic properties
could then be extracted). Thus although I will ground the discussion
of the TreeBanker in its use in adapting the CLE system to the ATIS
domain, the work described is of much more general application.

\section{Discriminant-Based Training}
\label{interact}

Many of the properties extracted from QLFs can be presented to
non-expert users in a form they can easily understand.
Those properties that hold for some analyses of a particular utterance
but not for others I will refer to as {\it discriminants} (Dagan and
Itai, 1994; Yarowsky, 1994). Discriminants that fairly consistently
hold for correct but not (some) incorrect analyses, or vice versa, are
likely to be useful in distinguishing correct from incorrect analyses
at run time. Thus for training on an utterance to be effective, we
need to provide enough ``user-friendly'' discriminants to allow the
user to select the correct analyses, and as many as possible
``system-friendly'' discriminants that, over the corpus as a whole,
distinguish reliably between correct and incorrect analyses. Ideally,
a discriminant will be both user-friendly and system-friendly, but
this is not essential. In the rest of this paper we will only
encounter user-friendly properties and discriminants.

\begin{figure*}
\begin{center}
\setlength{\unitlength}{1in}
\begin{picture}(6.5,3.0)
\includegraphics{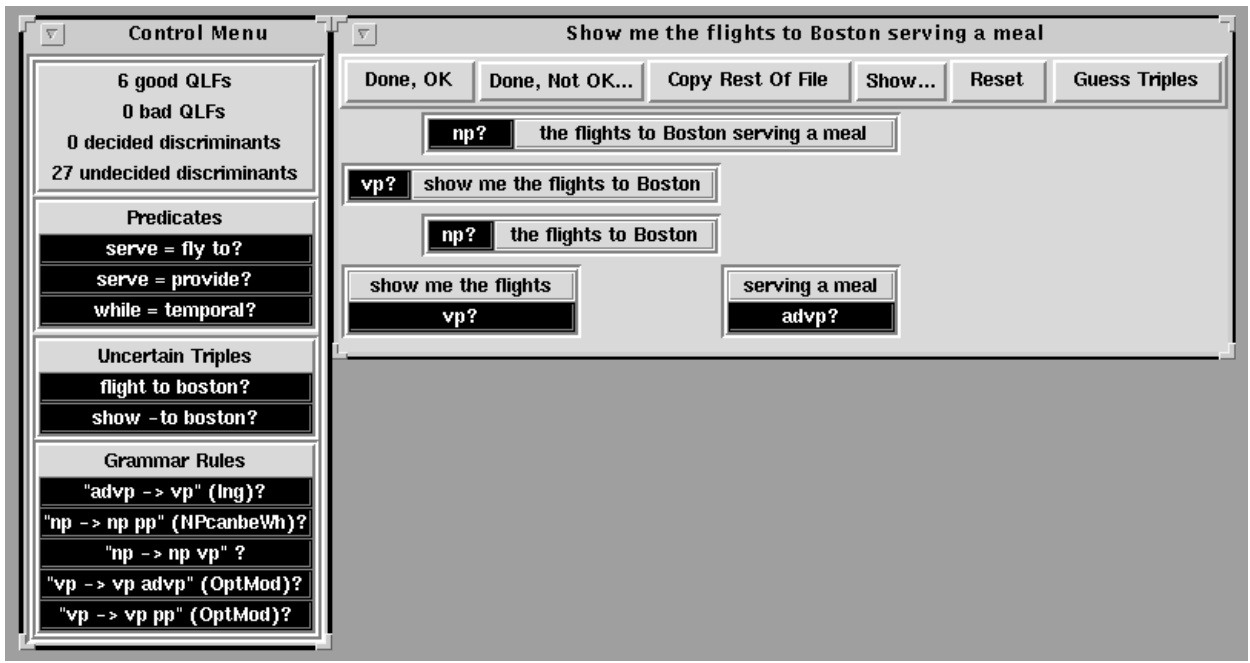}
\end{picture}
\caption{Initial TreeBanker display for ``Show me the flights to Boston
serving a meal''}
\label{snapshot1}
\end{center}
\end{figure*}

The TreeBanker presents properties to the user in a convenient
graphical form, exemplified in Figure \ref{snapshot1} for the sentence
``Show me the flights to Boston serving a meal''.
Initially, all discriminants are displayed in inverse video to show they
are viewed as undecided. Through the disambiguation process,
discriminants and the analyses they apply to can be undecided, correct
(``good'', shown in normal video), or incorrect (``bad'', normal video
but preceded a negation symbol ``\verb!~!'').  The user may click on any
discriminant with the left mouse button to select it as correct, or
with the right button to select it as incorrect. The types of property
currently extracted, ordered approximately from most to least
user-friendly, are as follows; examples are taken from the six QLFs
for the sentence used in figure \ref{snapshot1}.

\begin{itemize}
\item {\it Constituents}: ADVP for ``serving a meal'' (a discriminant,
holding only for readings that could be paraphrased ``show me the
flights to Boston while you're serving a meal''); VP for ``serving
a meal'' (holds for all readings, so not a discriminant and not shown
in figure \ref{snapshot1}).
\item {\it Semantic triples}: relations between word senses mediated
usually by an argument position, preposition or conjunction (Alshawi
and Carter, 1994).  Examples here (abstracting from senses to root
word forms, which is how they are presented to the user) are ``{\tt
flight to Boston}'' and ``{\tt show -to Boston}'' (the ``{\tt -}''
indicates that the attachment is not a low one; this distinction is
useful at run time as it significantly affects the likelihood of such
discriminants being correct). Argument-position relations are less
user-friendly and so are not displayed.

When used at run time, semantic triples undergo abstraction to a set
of semantic classes defined on word senses. For example, the obvious
senses of ``Boston'', ``New York'' and so on all map onto the class
name {\tt cc\_city}. These classes are currently defined manually by
experts; however, only one level of abstraction, rather than a full
semantic hierarchy, seems to be required, so the task is not too
arduous.
\item {\it Word senses}: ``serve'' in the sense of ``fly to'' (``does
United serve Dallas?'') or ``provide'' (``does that flight serve
meals?'').
\item {\it Sentence type}: imperative sentence in this case
(other moods are possible; fragmentary sentences are displayed as
``elliptical NP'', etc).
\item {\it Grammar rules used}: the rule name is given. This can be
useful for experts in the minority of cases where their intervention
is required.
\end{itemize}

In all, 27 discriminants are created for this sentence, of which 15
are user-friendly enough to display, and a further 28 non-discriminant
properties may be inspected if desired.  This is far more than the
three distinct differences between the analyses (``serve'' as ``fly
to'' or ``provide''; ``to Boston'' attaching to ``show'' or
``flights''; and, if ``to Boston'' does attach to ``flights'', a
choice between ``serving a meal'' as relative or adverbial). The
effect of this is that the user can give attention to whatever
discriminants he\footnote{I make the customary apologies for this use
of pronouns, and offer the excuse that most use of the TreeBanker to
date has been by men.} finds it easiest to judge; other, harder ones
will typically be resolved automatically by the TreeBanker as it
reasons about what combinations of discriminants apply to which
analyses. The first rule the TreeBanker uses in this reasoning process
to propagate decisions is:
\begin{itemize}
\item[R1] If an analysis (represented as a set of discriminants) has a
discriminant that the user has marked as bad, then the analysis
must be bad.
\end{itemize}
This rule is true by definition. The other rules used depend on the
assumption that there is exactly one good analysis among those that
have been found, which is of course not true for all sentences; see
Section \ref{additional} below for the ramifications of this.
\begin{itemize}
\item[R2] If a discriminant is marked as good, then only analyses of
which it is true can be good (since there is at most one good analysis).
\item[R3] If a discriminant is true only of bad analyses, then it is
bad (since there is at least one good analysis).
\item[R4] If a discriminant is true of all the undecided analyses,
then it is good (since it must be true of the correct one, whichever
it is).
\end{itemize}

\begin{figure*}
\begin{center}
\setlength{\unitlength}{1in}
\begin{picture}(6.5,3.0)
\includegraphics{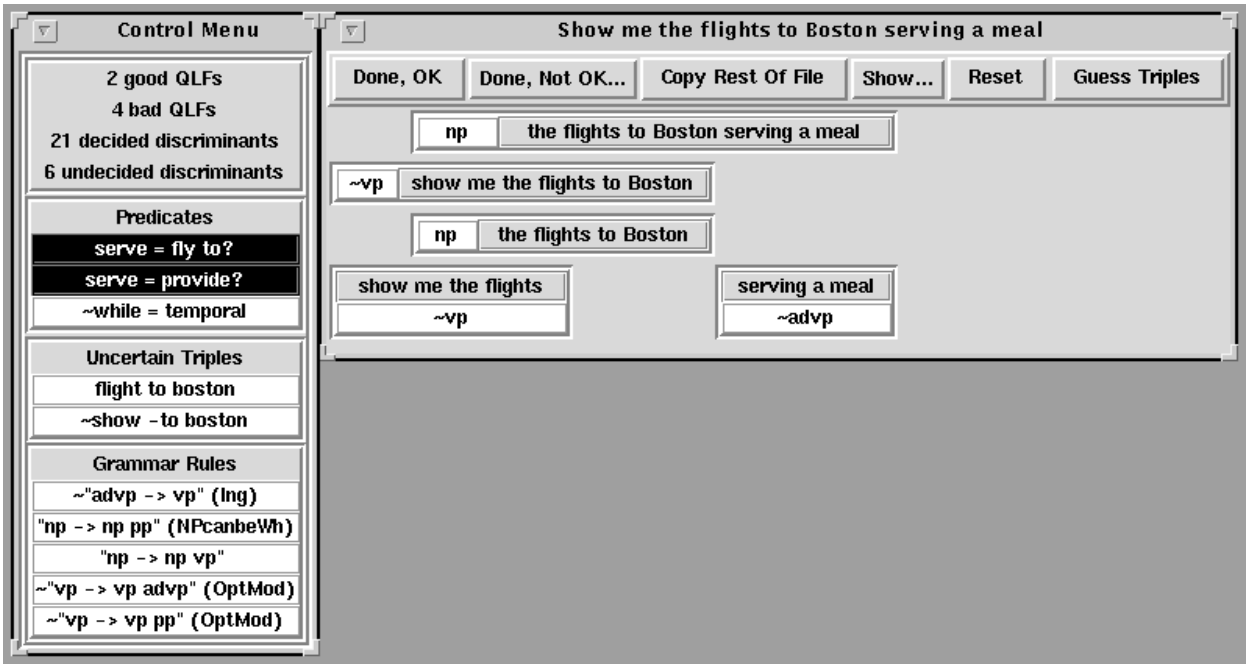}
\end{picture}
\caption{TreeBanker display after approving topmost ``np'' discriminant}
\label{snapshot2}
\end{center}
\end{figure*}

Thus if the user selects ``the flights to Boston serving a meal'' as
a correct NP, the TreeBanker applies rule R2 to narrow down the set of
possible good analyses to just two of the six (hence the item ``2 good
QLFs'' at the top of the control menu in the figure; this is really a
shorthand for ``2 possibly good QLFs''). It then applies R1-R4
to resolve {\it all} the other discriminants except the two for the
sense of ``serve''; and only those two remain highlighted in inverse
video in the display, as shown in Figure \ref{snapshot2}.
So, for example, there is no need for the user explicitly to make the trickier
decision about whether or not ``serving a meal'' is an adverbial
phrase. The user simply clicks on ``{\tt serve = provide}'', at which point
R2 is used to rule out the other remaining analysis and then R3 to
decide that ``{\tt serve = fly to}'' is bad.

The TreeBanker's propagation rules often act like this to simplify
the judging of sentences whose discriminants combine to produce an
otherwise unmanageably large number of QLFs.  As a further example,
the sentence ``What is the earliest flight that has no stops from
Washington to San Francisco on Friday?'' yields 154 QLFs and 318
discriminants, yet the correct analysis may be obtained with only two
selections.  Selecting ``the earliest flight ... on Friday'' as an NP
eliminates all but twenty of the analyses produced, and approving
``that has no stops'' as a relative clause eliminates eighteen of
these, leaving two analyses which are both correct for the purposes of
translation.  152 incorrect analyses may thus be dismissed in less
than fifteen seconds.

\begin{figure*}
\begin{center}
\setlength{\unitlength}{1in}
\begin{picture}(6.5,4.18)
\includegraphics{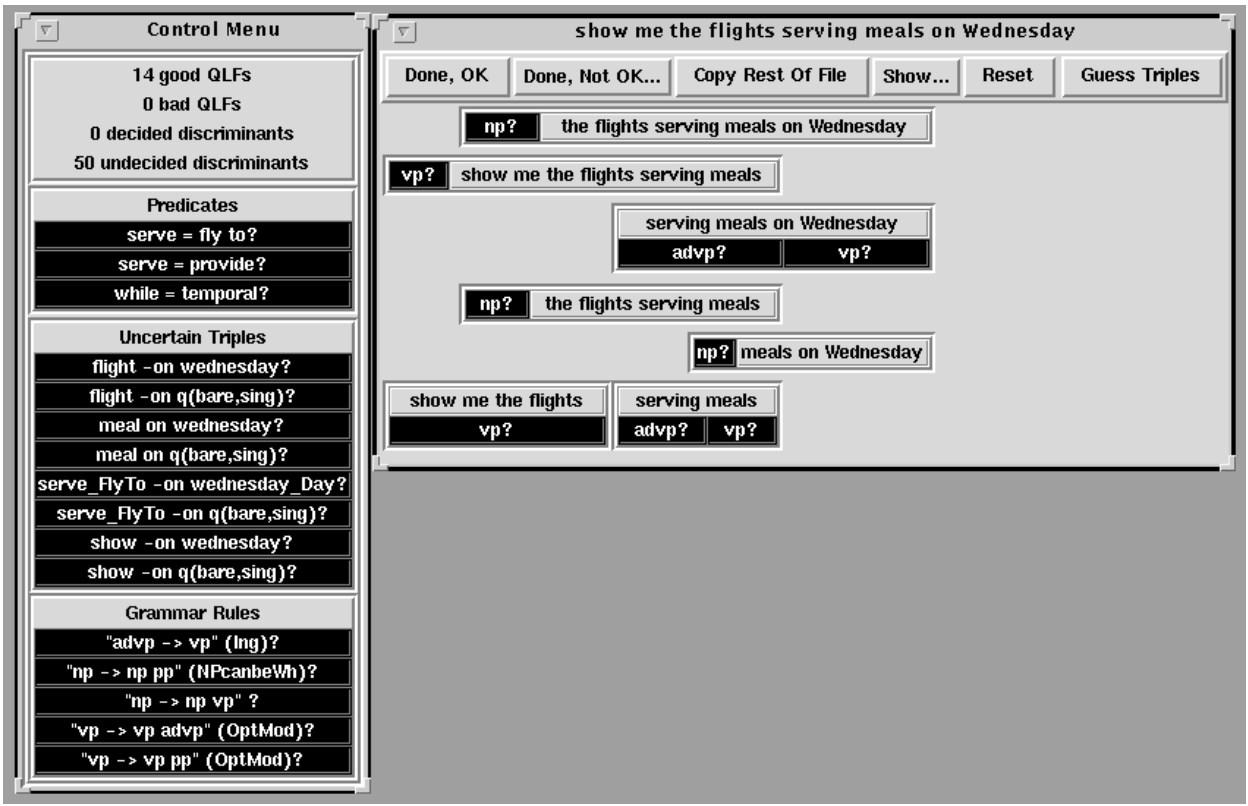}
\end{picture}
\caption{Initial TreeBanker display for ``Show me the flights serving
meals on Wednesday''}
\label{snapshot3}
\end{center}
\end{figure*}

The utterance ``Show me the flights serving meals on Wednesday''
demonstrates the TreeBanker's facility for presenting the user with
multiple alternatives for determining correct analyses.  As shown in
Figure \ref{snapshot3}, the following decisions must be made:
\begin{itemize}
\item Does ``serving'' mean ``flying to'' or ``providing''?
\item Does ``on Wednesday''
modify ``show'', ``flights'', ``serving'' or ``meals''?
\item Does ``serving'' modify ``show'' or ``flights''?
\end{itemize}
but this can be done by approving and rejecting various constituents
such as ``the flights serving meals'' and ``meals on Wednesday'', or
through the selection of triples such as ``flight -on Wednesday''.
Whichever method is used, the user can choose among the 14 QLFs
produced for this sentence within twenty seconds.

\section{Additional Functionality}
\label{additional}

Although primarily intended for the disambiguation of corpus sentences
that are within coverage, the TreeBanker also supports the diagnosis
and categorization of coverage failures. Sometimes, the user may
suspect that {\it none} of the provided analyses for a sentence is
correct. This situation often becomes apparent when the TreeBanker
(mis-)applies rules R2-R4 above and insists on automatically assigning
incorrect values to some discriminants when the user makes decisions
on others; the coverage failure may be confirmed, if the user is
relatively accomplished, by inspecting the non-discriminant properties
as well (thus turning the constituent window into a display of the
entire parse forest) and verifying that the correct parse tree is not
among those offered.  Then the user may mark the sentence as ``Not
OK'' and classify it under one of a number of failure types,
optionally typing a comment as well. At a later stage, a system expert
may ask the TreeBanker to print out all the coverage failures of a
given type as an aid to organizing work on grammar and lexicon
development.

For some long sentences with many different readings, more
discriminants may be displayed than will fit onto the screen at one
time. In this case, the user may judge one or two discriminants
(scrolling if necessary to find likely candidates), and ask the
TreeBanker thereafter to display only {\it undecided} discriminants;
these will rapidly reduce in number as decisions are made, and can
quite soon all be viewed at once.

If the user changes his mind about a discriminant, he can click on it
again, and the TreeBanker will take later judgments as superceding
earlier ones, inferring other changes on that basis. Alternatively,
the ``Reset'' button may be pressed to undo all judgments for the
current sentence.

It has proved most convenient to organize the corpus into files that
each contain data for a few dozen sentences; this is enough to
represent a good-sized corpus in a few hundred files, but not so big
that the user is likely to want to finish his session in the middle of
a file.

Once part of the corpus has been judged and the information extracted
for run-time use (not discussed here), the TreeBanker may be told to
resolve discriminants automatically when their values can safely be
inferred. In the ATIS domain, ``{\tt show \verb!-to! (city)}'' is a triple that is
practically never correct, since it only arises from incorrect PP
attachments in sentences like ``Show me flights to New York''. The
user can then be presented with an initial screen in which that
choice, and others resulting from it, are already made. This speeds up
his work, and may in fact mean that some sentences do not need to be
presented at all.

In practice, coverage development tends to overlap somewhat with the
judging of a corpus. In view of this, the TreeBanker includes a
``merge'' option which allows existing judgments applying to an old
set of analyses of a sentence to be transferred to a new set that
reflects a coverage change. Properties tend to be preserved much
better than whole analyses as coverage changes; and since only
properties, and not analyses, are kept in the corpus database, the
vast bulk of the judgments made by the user can be preserved.

The TreeBanker can also interact directly with the CLE's analysis
component to allow a user or developer to type sentences to the
system, see what discriminants they produce, and select one analysis
for further processing. This configuration can be used in a number of
ways. Newcomers can use it to familiarize themselves with the system's
grammar. More generally, beginning students of grammar can use it to
develop some understanding of what grammatical analysis involves.  It
is also possible to use this mode during grammar development as an aid
to visualizing the effect of particular changes to the grammar on
particular sentences.

\section{Evaluation and Conclusions}
\label{evalconc}

Using the TreeBanker, it is possible for a linguistically aware
non-expert to judge around 40 sentences per hour after a few days
practice. When the user becomes still more practised, as will be the
case if he judges a corpus of thousands of sentences, this figure
rises to around 170 sentences per hour in the case of our most
experienced user. Thus it is reasonable to expect a corpus of 20,000
sentences to be judged in around three person weeks.  A much smaller
amount of time needs to be spent by experts in making judgments he
felt unable to make (perhaps for one per cent of sentences once the
user has got used to the system) and in checking the user's work (the
TreeBanker includes a facility for picking out sentences where errors
are mostly likely to have been made, by searching for discriminants
with unusual values). From these figures it would seem that the
TreeBanker provides a much quicker and less skill-intensive way to
arrive at a disambiguated set of analyses for a corpus than the manual
annotation scheme involved in creating the Penn Treebank; however, the
TreeBanker method depends on the prior existence of a grammar for the
domain in question, which is of course a non-trivial requirement.

Engelson and Dagan (1996) present a scheme for selecting corpus
sentences whose judging is likely to provide useful new information,
rather than those that merely repeat old patterns. The TreeBanker
offers a related facility whereby judgments on one sentence may be
propagated to others having the same sequence of parts of speech.
This can be combined with the use of {\it representative corpora} in
the CLE (Rayner, Bouillon and Carter, 1995) to allow only one
representative of a particular pattern, out of perhaps dozens in the
corpus as a whole, to be inspected. This already significantly reduces
the number of sentences needing to be judged, and hence the time
required, and we expect further reductions as Engelson's and Dagan's
ideas are applied at a finer level.

In the current implementation, the TreeBanker only makes use of {\it
context-independent} properties: those derived from analyses of an
utterance that are constructed without any reference to the context of
use. But utterance disambiguation in general requires the use of
information from the context. The context can influence choices of
word sense, syntactic structure and, most obviously, anaphoric reference
(see e.g.\ Carter, 1987, for an overview), so it might seem that a
disambiguation component trained only on context-independent
properties cannot give adequate performance.

However, for QLFs for the ATIS domain, and presumably for others of
similar complexity, this is not in practice a problem. As explained
earlier, anaphors are left unresolved at the stage of analysis and
disambiguation we are discussing here; and contextual factors for
sense and structural ambiguity resolution are virtually always
``frozen'' by the constraints imposed by the domain. For example,
although there are certainly contexts in which ``Tell me flights to
Atlanta on Wednesday'' could mean ``Wait until Wednesday, and then
tell me flights to Atlanta'', in the ATIS domain this reading is
impossible and so ``on Wednesday'' must attach to ``flights''.  For a
wider domain such as NAB, one could perhaps attack the context problem
either by an initial phase of topic-spotting (using a different set of
discriminant scores for each topic category), or by including some
discriminants for features of the context itself among these to which
training was applied.

\section*{Acknowledgements}

I am very grateful to Martin Keegan for feedback on his hard work of
judging 16,000 sentences using the TreeBanker, and to Manny Rayner,
David Milward and anonymous referees for useful comments on earlier
versions of this paper.

The work reported here was begun under funding from by the Defence
Research Agency, Malvern, UK, under Strategic Research Project
AS04BP44, and continued with funding from Telia Research AB under the
SLT-2 project.

\section*{References}

\parskip 4pt

\noindent
Becket, Ralph, and 19 others (forthcoming). {\em Spoken Language
Translator: Phase Two Report}. Joint report by SRI International and
Telia Research.

\noindent
Alshawi, Hiyan, editor (1992). {\em The Core Language Engine}.  The
MIT Press, Cambridge, Massachusetts.

\noindent
Alshawi, Hiyan, and David Carter (1994). ``Training and Scaling
Preference Functions for Disambiguation''. {\it Computational
Linguistics}, 20:4.*\footnote{Starred references are also available
from \verb!http://www.cam.sri.com!.}

\noindent
Alshawi, Hiyan, and Richard Crouch (1992).  ``Monotonic Semantic
Interpretation''.  In {\em Proceedings of 30th Annual Meeting of the
Association for Computational Linguistics}, pp.~32--39, Newark,
Delaware.*

\noindent
Carter, David (1987). ``Interpreting Anaphors in Natural Language
Texts''. Chichester: Ellis Horwood.

\noindent
Dagan, Ido, and Alon Itai (1994). ``Word Sense Disambiguation Using a
Second Language Monolingual Corpus'', {\it Computational Linguistics}
20:4, pp.~563--596.

\noindent
Engelson, Sean, and Ido Dagan (1996). ``Minimizing Manual Annotation
Cost in Supervised Training from Corpora''. In {\em Proceedings of 34th Annual
Meeting of the Association for Computational Linguistics},
pp.~319-326, Santa Cruz, CA.

\noindent
Gamb\"{a}ck, Bj\"{o}rn, and Manny Rayner (1992). ``The Swedish Core
Language Engine''.  In {\em Proceedings of NOTEX-92}.*

\noindent
Hemphill,~C.T., J.J.\ Godfrey and G.R.\ Doddington (1990).
``The ATIS Spoken Language Systems pilot corpus.''
Proceedings of DARPA Speech and Natural Language Workshop, Hidden
Valley, Pa., pp.\ 96-101.

\noindent
Marcus, Mitchell, Beatrice Santorini, and Mary Ann Marcinkiewicz
(1993). ``Building a Large Annotated Corpus of English: the Penn
Treebank''. {\it Computational Linguistics}, 19:2, pp.~313-330.

\noindent
Murveit, Hy, John Butzberger, Vassilios Digalakis and Mitchell
Weintraub (1993). ``Large Vocabulary Dictation using SRI's
DECIPHER(TM) Speech Recognition System: Progressive Search
Techniques''. In {\em Proceedings of ICASSP-93}.

\noindent
Rayner, Manny, Pierrette Bouillon, and David Carter (1995).  ``Using
Corpora to Develop Limited-Domain Speech Translation Systems''. In
{\em Proceedings of Translating and the Computer 17}, ASLIB, London.*

\noindent
Rayner, Manny, David Carter, and Pierrette Bouillon (1996).
``Adapting the Core Language Engine to French and Spanish''.  In {\em
Proceedings of NLP-IA}, Moncton, New Brunswick.*


\noindent
Rayner, Manny, and David Carter (1996). ``Fast Parsing using Pruning
and Grammar Specialization''. In {\em Proceedings of 34th Annual
Meeting of the Association for Computational Linguistics},
pp.~223--230, Santa Cruz, CA.*

\noindent
Rayner, Manny, and David Carter (1997). ``Hybrid language processing
in the Spoken Language Translator''. In {\em Proceedings of
ICASSP-97}.*

\noindent
Yarowsky, David (1994). ``Decision Lists for Lexical Ambiguity
Resolution''.  In {\em Proceedings of 32nd Annual Meeting of the
Association for Computational Linguistics}, pp.~88-95, Las Cruces, NM.

\end{document}